\documentclass[conference]{IEEEtran}
\IEEEoverridecommandlockouts
\usepackage{cite}
\usepackage{amsmath,amssymb,amsfonts}
\usepackage{algorithmic}
\usepackage{graphicx}
\usepackage{textcomp}
\usepackage{xcolor}
\usepackage[utf8]{inputenc}
\usepackage{float}

\def\BibTeX{{\rm B\kern-.05em{\sc i\kern-.025em b}\kern-.08em
    T\kern-.1667em\lower.7ex\hbox{E}\kern-.125emX}}
\begin{document}

\title{Curriculum Vitae Recommendation\\ Based on Text Mining}

\author{\IEEEauthorblockN{Honorio Apaza Alanoca}
\IEEEauthorblockA{Data Science Research Group \\
National University of Moquegua\\
Ilo, Moquegua, Perú\\
honorio.apz@gmail.com}
\and
\IEEEauthorblockN{Américo A. Rubin de Celis Vidal}
\IEEEauthorblockA{Informatics Department \\
National University of Moquegua\\
Ilo, Moquegua, Perú\\
americorubin@gmail.com}
\and
\IEEEauthorblockN{Josimar Edinson Chire Saire}
\IEEEauthorblockA{Institute of Mathematics and\\
Computer Science (ICMC) \\
University of São Paulo (USP)\\
São Carlos, SP, Brazil\\
jecs89@usp.br}
}

\maketitle

\begin{abstract}

During the last years, the development in diverse areas related to computer science and internet, allowed to generate new alternatives for decision making in the selection of personnel for state and private companies. In order to optimize this selection process, the recommendation systems are the most suitable for working with explicit information related to the likes and dislikes of employers or end users, since this information allows to generate lists of recommendations based on collaboration or similarity of content. Therefore, this research takes as a basis these characteristics contained in the database of curricula and job offers, which correspond to the Peruvian ambit, which highlights the experience, knowledge and skills of each candidate, which are described in textual terms or words. 
This research focuses on the problem: how we can take advantage from the growth of unstructured information about job offers and curriculum vitae on different websites for CV recommendation. So, we use the techniques from Text Mining and Natural Language Processing. Then, as a relevant technique for the present study, we emphasize the technique frequency of the Term - Inverse Frequency of the documents
(TF-IDF), which allows identifying the most relevant CVs in relation to a job offer of website through the average values (TF-IDF). So,  the weighted value can be used as a qualification value of the relevant curriculum vitae for the recommendation.
\end{abstract}

\begin{IEEEkeywords}
Natural Language Processing, Text Mining, System Recommender, Job Offer
\end{IEEEkeywords}

\section{Introduction}

In concordance with the report of Internet World Stats 2019\cite{b1}, the access to the Internet and web technology has grown fastly and let the users interact trough different platforms. 
Considering the ambit related to the working  market, and the high demand and offer of job places, it is possible to observe that  there is a set of unstructured data which contain important information related to the profile of candidates for a particular job, some of these websites as: Computrabajo, Bumeran, Info Jobs, etc., thatusually has the profile of the positions and the Curriculum Vitae(CV) of the candidates.

Frequently, the users have specific preferences\cite{b2} related with some products or objects, these preferences are present in non-explicit text then the information about the likes or dislikes must be extracted from the text. The preferences of one user usually are represented using a matrix, and the content of the matrix has the level of preference for one specific product.
Thus, the research on Text Mining(TM)\cite{b3} uses Natural Language Processing(NLP) to extract information from the text written by human beings. The automatization is necessary for the big number of sources( emails, documents, social networks, etc.)

Recommendation Systems(RS) are useful to suggest or recommend one item( product, object, etc.) to one user based on the information from other users. Usually, RS works with text then NLP algorithms\cite{b4,b5,b6} are used to extract information, relevant terms from the text source.

The present work explains the process step by step to build a Job Recommendation based on Web Scrapping, NLP an TM algorithms.
Chapter II includes the review of the bibliography, chapter III discusses the methodology to be used, chapter IV develops the work proposal, chapter V discloses the results of the research and finally in Chapter VI gives conclusions.
\section{Literature's Review}
The study Employers expectations: A probabilistic text mining model \cite{b7}, more than 20,000 job advertisements from various websites were processed, the method of text mining was applied to identify information skills derived from the web pages of the construction industry sector.
In the research named Text Analysis for Job Matching Quality Improvement \cite{b5}, in a context of data analysis that includes travel time, job location, job type, rates, candidate skill set, etc. And when applying keywords in a machine learning process using text mining tools, as a result, effective keywords are discovered for a job matching system.
in the research entitled Natural Language Processing and Text Mining to Identify Knowledge Profiles for Software Engineering Positions \cite{b8}, through the application of NLP and TM to analyze the unstructured text of the resumes and job offers, it manages to identify the knowledge profiles for software engineering positions.
In the research entitled Data Mining Approach to Monitoring The Requirements of the Job Market: A Case Study \cite{b9}, presents an approach based on data mining to identify the most demanded occupations in the modern labor market. To achieve this, have a latent semantic indexing model that is able to match the job announcement extracted from the 18web with the data of the occupation description in the database.
\section{Methodology}
\begin{figure}[htbp]
\centerline{\includegraphics [width=0.4\textwidth]{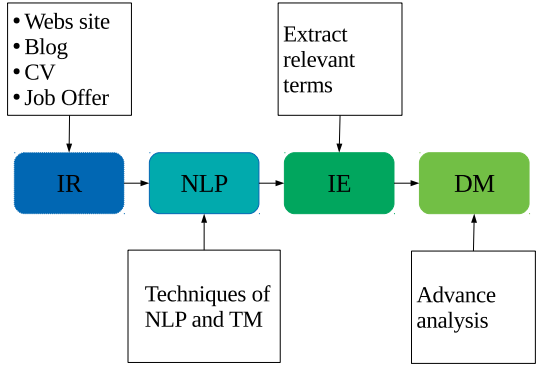}}
\caption{Methodology flowchart.}
\label{fig:fisrt}
\end{figure}

In general terms the text mining is a process to turn the text unstructured in data structured for analysis, to achieve that, is necessary applied algorithms artificial intelligence and statistics techniques  to text documents. That process can be combined in a workflow \cite{b10} \cite{b11}, as can see in the figure \eqref{fig:fisrt}.

\begin{itemize}
    \item Information retrieval(RI), Collect data about one topic or many from the text sources(websites, emails, etc.).
    \item Natural Language Processing(NLP), Interpretation of the text considering linguistics supported by Machine Learning techniques.
    \item Information Extraction(IE), Identification of relevant terms for a summarized representation of documents\cite{b12}.
    \item Data Mining(DM), Exploitation of structured data to discover knowledge about relationships.
\end{itemize}
\section{Proposal}
\subsection{Information  retrieval}
Initially was apply Scrapy to recover data in Spanish language about job offers and Curriculum Vitae (CVs) of the candidates from different websites as: computrabajo, Bumeran etc. for the purpose of research.  
A collection a set of 10,000 job offers, with the next format: title, city, company's name, description, requirements, functions and previous knowledge for the job role.
The set of 10,000 Curriculum Vitae was gathered in HTML code, with the next format: name, description, skills and experience, and that data was extracted thanks to BeautifulSoup library. 

\subsection{Natural Language Processing}
\subsubsection{Cleaning and Transformation of Data}
Initially it was used the package NLTK (Natural Language Toolkit). Next, each of these data collected are processed by the functions:
\begin{itemize}
\item remove\_non\_ascii\_(): For limitation of no-ASCII characters.
\item to\_lowercase\_(): For data normalization to lowecase.
\item remove\_punctuation\_(): To remove scores and accents of Spanish language.
\item remove\_stopwords\_(): To remove empty words from the Spanish language (el, la, los, en, etc.).
\item lemmatize\_verbs\_(): For the randomization of data verbs.
\end{itemize}
These function set  allow obtain  clean and standardized data, at this stage you are ready to apply TF-IDF (Term Frequency - Inverse Document Frequency) algorithm.
The entry information is structured in two general lists, (resumes and job offers).

\subsubsection{Term Frequency-Inverse Document Frequency(TF-IDF)}
TF-IDF represents a numerical relevance of one term for one document and set of documents(corpus).

\begin{equation}
tf\_idf_{t,d} = tf_{t,d} \cdot \log \frac{N}{df_t}\label{eq:one}
\end{equation}
Where:\\
$tf\_idf$ = Term Frequency-Inverse Document. Frequency\\
$tf_{t,d}$ = Occurrences Frequency of term $t$ in document $d$.\\
$N$ = Total Number of Documents.\\
$df_{t}$ = Number of documents with the term $t$.\\

\subsubsection{Information distribution}
The data collected on CVs are divided into three groups: description (D1), experience (D2) and skills(D3) respectively.\\
The information collected on job offers are divided into five groups of terms, such as: Title, description, requirements, functions and knowledge. These groups are united in one only after applying NLTK, in addition the duplicate terms are removed to avoid redundancy when measuring the relevance of these terms in the CVs.
\subsubsection{Application TF-IDF}
To find the relevance value of each of the terms of the job offers in the curriculum vitae documents, the technique frequency of terms - inverse frequency of documents (TF-IDF) was applied.\\
Note that, the  equation \eqref{eq:one} of the TF-IDF algorithm consists of two important parts, the first is to find the frequency term value ($tf$) in a given document, the second is to find the inverse frequency value of $N$ documents ($idf$). 

The use of TF-IDF for filtering terms in documents for content-based recommendation system is used and recommended in the investigation of \cite{b13}.
In this paper, the TF-IDF technique is applied to measure the relevance of the terms of job offers in resumes, the weighted value is an indicator of relevance of a CV with respect to job offers.\\
The relevance of the terms of the job offers are calculated in terms of CV, the terms are distributed in three parts, called documents (description, experience and skills). \\
Table \eqref{tab1} shows the example of the five terms, where: the fields TF-D1, TF-D2 and TF-D3 correspond to the calculation of frequency ($tf$) of terms of job offers in respective CV documents. The IDF field corresponds to calculation of inverse frequency  of documents ($idf$), and in the TF-IDF-D1, TF-IDF-D2 and TF-IDF-D3 fields correspond to the $TF-IDF$ calculation for respective documents.
Next, the tf-idf values are weighted for prescriptive documents, and finally the general average of all the documents is obtained.
\begin{table*}[htbp]
\caption{Term relevance calculation process}
\begin{center}
\begin{tabular}{|c|c|c|c|c|c|c|c|c|}
\hline
\textbf{\textit{Terms}}& \textbf{\textit{TF-D1}}&\textbf{\textit{TF-D2}}&\textbf{\textit{TF-D3}}&\textbf{\textit{IDF}}&\textbf{\textit{TF-IDF-D1}}&\textbf{\textit{TF-IDF-D2}}& \textbf{\textit{TF-IDF-D3}} \\
\hline
Minería & 2/21 & 1/5 & 0/5 & Log(3/2) & 0.039 & 0.081 & 0  \\
Scrapy & 1/21 & 0/5 & 1/5 & Log(3/2) & 0.019 & 0 & 0.081  \\
Aplicando & 1/21 & 0/5 & 0/5 & Log(3/1) & 0.052 & 0 & 0  \\
Sistema & 2/21 & 1/5 & 0/5 & Log(3/2) & 0.039 & 0.081 & 0  \\
Recomendación & 2/21 & 1/5 & 0/5 & Log(3/2) & 0.039 & 0.081 & 0  \\
$n$ text & ... & ... & ... & ... & ... & ... & ...  \\
\hline
\multicolumn{5}{l}{Average TF-IDF for documents:}&{0.04}&{0.109}&{0.127}\\
\hline
\multicolumn{7}{l}{general average TF-IDF of the documents:}&{0.092}\\
\hline
\end{tabular}
\label{tab1}
\end{center}
\end{table*}

\subsection{Information Extraction}
Table \eqref{tab2} show the averages of $TF-IDF$ detailed by document and last field show a average general, for this work could be useful for the recommendation of job offers and other data mining purposes. So that information was extracted to use in different contexts, on this case for CV recommendation.
\begin{table}[htbp]
\caption{Relevance of CV with respect to the job offer.}
\begin{center}
\begin{tabular}{|c|c|c|c|c|c|}
\hline
\textbf{\textit{Idjob}}& \textbf{\textit{Idcv}}& \textbf{\textit{Rating}}& \textbf{\textit{Rating}}& \textbf{\textit{Rating}}& \textbf{\textit{Averages}} \\
\textbf{\textit{ }}& \textbf{\textit{ }}& \textbf{\textit{Description}}& \textbf{\textit{Experience}}& \textbf{\textit{Skills}}& \textbf{\textit{General}} \\
\hline
0& 0& 0.04& 0.109& 0.127&  0.092 \\
0& 1& 0 & 0& 0 & 0\\
0& 2& 0.005& 0.006& 0 & 0.004 \\
0& 3& 0.002& 0.003& 0 & 0.002\\
0& 4& 0& 0& 0.037 & 0.012 \\
\hline
\end{tabular}
\label{tab2}
\end{center}
\end{table}

\subsection{Data mining}
On the results of the previous stage, we can analyze and visualize the usefulness of this data, the results can help different cases. In the Table \eqref{tab3}, we can see a matrix of 5000 resumes for 9998 jobs, where the relevance value of each curriculum vitae is distributed on a matrix basis.
\begin{table}[htbp]
\caption{Data mining about job offers and CV.}
\begin{center}
\begin{tabular}{|c|c|c|c|c|c|c|c|c|}
\hline
\textbf{\textit{Id Cv}}& \textbf{\textit{0}}& \textbf{\textit{1}}& \textbf{\textit{2}}& \textbf{\textit{...}}&  \textbf{\textit{4997}}& \textbf{\textit{4998}}& \textbf{\textit{4999}} \\
\textbf{\textit{Id Job}}& \textbf{\textit{ }}& \textbf{\textit{ }}& \textbf{\textit{ }}& \textbf{\textit{ }}& \textbf{\textit{ }}& \textbf{\textit{ }}& \textbf{\textit{ }} \\

\hline
0 & 0.092 & 0.000 & 0.004 & ... & 0.015 & 0.001 & 0.000 \\
1 & 0.000 & 0.008 & 0.013 & ... & 0.011 & 0.012 & 0.004 \\
2 & 0.000 & 0.023 & 0.013 & ... & 0.012 & 0.001 & 0.002 \\
3 & 0.000 & 0.023 & 0.013 & ... & 0.017 & 0.001 & 0.002 \\
4 & 0.000 & 0.025 & 0.013 & ... & 0.015 & 0.001 & 0.002 \\
5 & 0.000 & 0.022 & 0.013 & ... & 0.014 & 0.001 & 0.003 \\
... & ...& ... & ... & ... & ... & ... & ... \\
9997& 0.000 & 0.021 & 0.013 & ... & 0.015 & 0.000 & 0.000 \\
\hline
\end{tabular}
\label{tab3}
\end{center}
\end{table}
\\

\begin{figure}[htbp]
\centerline{\includegraphics [width=0.4\textwidth]{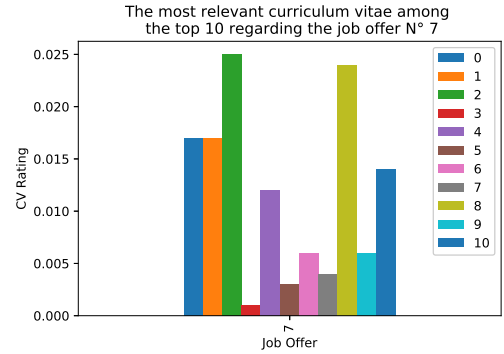}}
\caption{Visual relevance analysis of CV.}
\label{fig:two}
\end{figure}

With this information can do analysis about more relevant CVs with respect a one job offer, as be see in the Figure \eqref{fig:two}.  where it can clearly be seen that some Cvs are more relevant than others for the first job offer (Job7). The CVs 2nd and 8th are evidently more relevant for the 7ht job offer.
\\

\section{Results}
\begin{figure}[H]
\centerline{\includegraphics [width=0.5\textwidth]{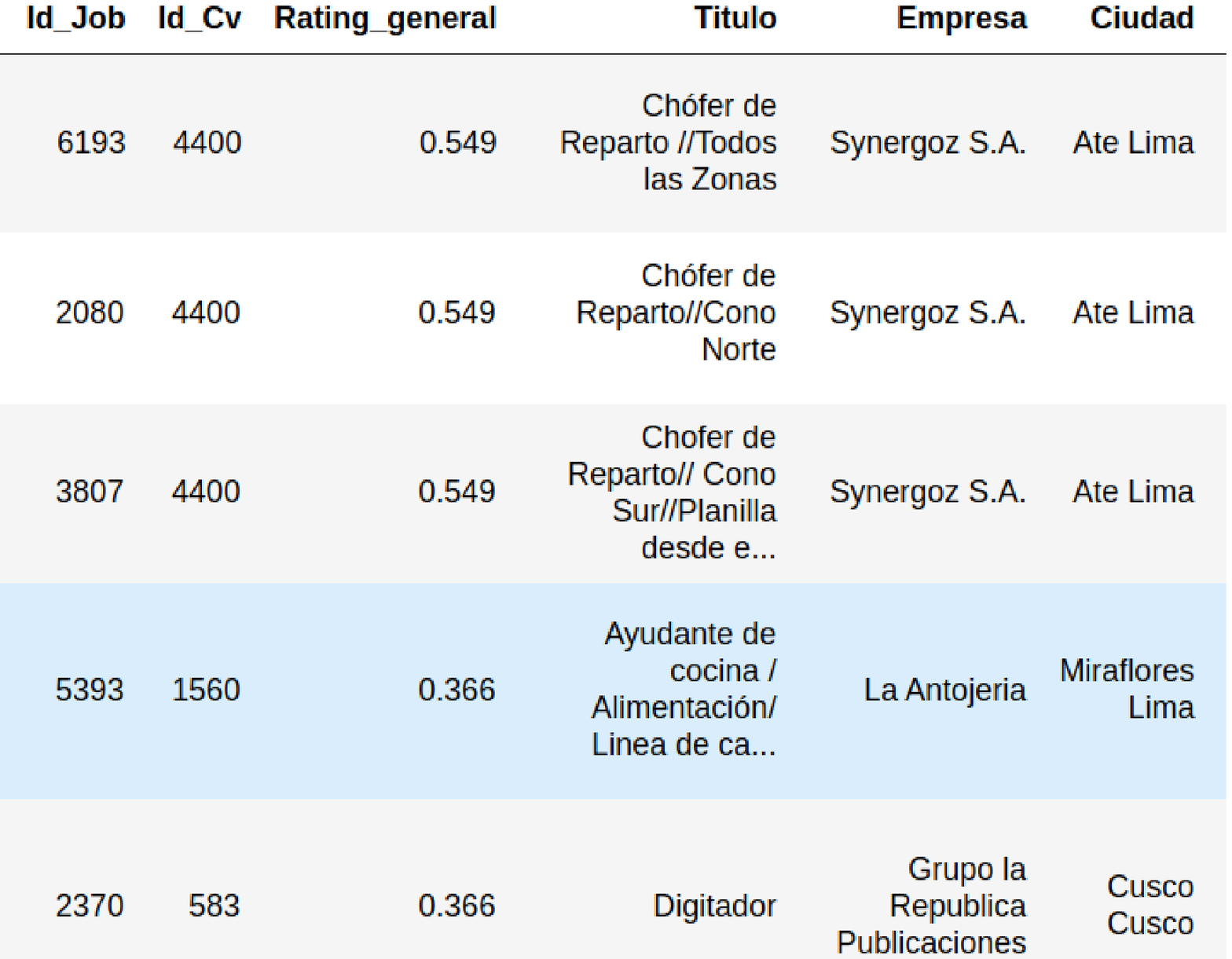}}
\caption{Data model with averages of relevance of CV with respect to job offers.}
\label{fig:thre}
\end{figure}

The present work provides a data model with averages of relevance of CV with respect to job offers, based on relevance of terms, that can be used to recommend CV or job offers, as shown in Figure \eqref{fig:thre}.\\
In the present case it is not necessary to test or validate results, because they are exact results based on relevance of terms, result of calculation of frequency of terms and inverse frequency of documents, therefore the results data are not probabilistic or approximations.

However, dataset is normalized base to max and min scaling, as can see in equation \eqref{eq:two}. when the value is approximate to 1 is most relevant and less relevant when it approximate to 0, in this dataset the highest value is 0.549 and the lowest 0.000.
\begin{equation}
    X_{cs}=\frac{X-X_{min}}{X_{max}-X_{min}}\label{eq:two}
\end{equation}

\section{Conclusions}
The results obtained from the CVs analyzed, clearly establish a set of classifications related to the relevance of each candidate in relation to the employer's job offer.

Based on these results, the employer can determine if in the geographic scope of his interest it is possible to find a candidate with the profile that the company requires.

The Term frequency - reverse document frequency ($ tf-idf $) proved to be effective in identifying more relevant information from a corpus of curriculum vitae data, the general averages of tf-idf could be considered as the rating value of the curriculum vitae for the recommendation of job offers.
\section{Recommendations}

The recommendation is continue the research on Recommendation Systems to take advantage of the proposal dataset with peruvian information about jobs. First, the employability of graduates from public/private universities is necessary and helpful to tune their programs up. Second, understand how important is the content of the courses offered to the students in university classrooms.
\\

\vspace{12pt}
\end{document}